\documentclass[runningheads]{llncs}

\usepackage{eccv}

\usepackage{eccvabbrv}

\usepackage{graphicx}
\usepackage{booktabs}

\usepackage[accsupp]{axessibility}  

\usepackage{hyperref}

\usepackage{orcidlink}
\usepackage{times}
\usepackage{epsfig}
\usepackage{graphicx}
\usepackage{amsmath}
\usepackage{amssymb}

\usepackage{booktabs}
\usepackage[utf8]{inputenc} 
\usepackage[T1]{fontenc}    
\usepackage{url}            
\usepackage{amsfonts}       
\usepackage{nicefrac}       
\usepackage{microtype}      
\usepackage{xcolor,colortbl}
\usepackage{multirow}
\usepackage{float}
\usepackage{comment}
\usepackage{caption}
\usepackage{subcaption}
\usepackage{soul}
\graphicspath{{images/}}

\begin{document}

\title{AA-SGAN: Adversarially Augmented Social GAN with Synthetic Data} 


\author{Mirko Zaffaroni\orcidlink{0000-0003-1111-4217} \and
Federico Signoretta\orcidlink{0000-0003-1365-2158} \and
Marco Grangetto\orcidlink{0000-0002-2709-7864} \and
Attilio Fiandrotti\orcidlink{0000-0002-9991-6822}}

\authorrunning{M.~Zaffaroni et al.}

\institute{Department of Computer Science, University of Turin, IT \\
\email{\{name.surname\}@unito.it}}

\maketitle

\begin{abstract}
  Accurately predicting pedestrian trajectories is crucial in applications such as autonomous driving or service robotics, to name a few. Deep generative models achieve top performance in this task, assuming enough labelled trajectories are available for training. To this end, large amounts of synthetically generated, labelled trajectories exist (e.g., generated by video games). However, such trajectories are not meant to represent pedestrian motion realistically and are ineffective at training a predictive model. We propose a method and an architecture to augment synthetic trajectories at training time and with an adversarial approach. We show that trajectory augmentation at training time unleashes significant gains when a state-of-the-art generative model is evaluated over real-world trajectories. Our source code and trained models are publicly available at: \href{https://github.com/mirkozaff/aa-sgan}{https://github.com/mirkozaff/aa-sgan}
\end{abstract}

\section{Introduction}
\label{sec:intro}

Predicting pedestrian trajectories is of paramount importance in applications where robots must dodge humans, e.g., to avoid collisions.
This task is inherently challenging due to the complexity and to some extent unpredictability of human movement patterns.
In detail, the task has three main technical challenges.
First, given a partial trajectory, multiple options for its continuation are possible, making the problem intrinsically multivariate;
second, an individual's motion is conditioned by bystanders' behaviours, especially in crowded spaces; finally, different social and cultural contexts may constrain what can be considered a plausible motion pattern.

\begin{figure}[t!]
  \centering
  \subfloat[Real trajectories]{\includegraphics[width=0.49\linewidth]{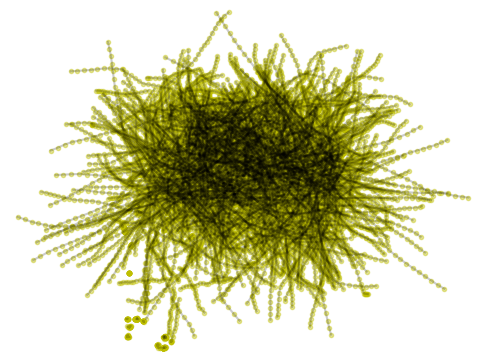}
  \label{fig:realplot}}
  \subfloat[Synthetic trajectories]{\includegraphics[width=0.49\linewidth]{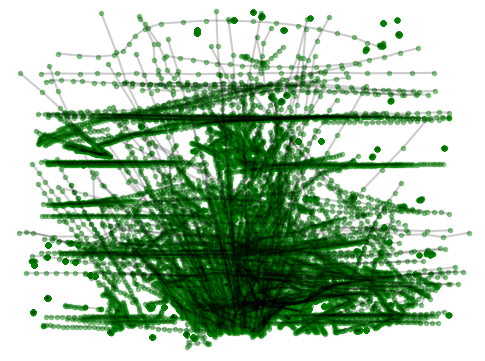}
  \label{fig:syntplot}}
  \vfill
  \subfloat[Synth-augmented trajectories]{\includegraphics[width=0.49\linewidth]{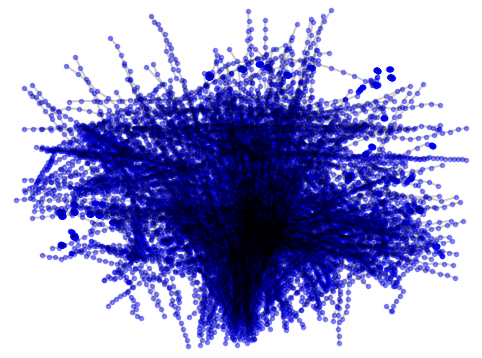}
  \label{fig:ganplot}}
  \caption[Real, synthetic, and synth-augmented  trajectories.] {Examples of real (a), synthetic (b), and synth-augmented (c) trajectories by the AA-SGAN Augmenter.
  }
  \label{fig:traj_comp}
\end{figure}

Recently, deep generative models showed promising results in plausible human motion prediction.
For example, Social GAN~\cite{gupta2018social} trains a trajectories Generator using an adversarial approach where ad-hoc architectural elements and loss terms promote trajectories that are \textit{socially} plausible.
This model is trained on pedestrian trajectories extracted from crowds footages from different environments~\cite{eth2009, ucy2007}; in Fig.~\ref{fig:realplot} sample trajectories extracted from a set of tracked pedestrians in a video scene are shown.
However, such datasets exhibit limitations such as no camera settings variability and yield comparatively few trajectories if compared to the capacity of some deep models.
While collecting more videos to enlarge the training set may bring benefits, data collection and annotation is a daunting, time-consuming, activity.
Also, releasing datasets of real footage poses some privacy-related issues.

Synthetic training allows for large training sets without the burden of manually collecting and labelling the samples.
For example, the Joint Track Auto Dataset (JTA)~\cite{fabbri2018learning} provides a large body of pedestrian trajectories from the videogame Grand Theft Auto V (GTAV).
However, such computer-generated trajectories lack the realism of real human motion: as an example in Fig.~\ref{fig:syntplot} many deterministic straight patterns present in the JTA dataset are visually evident.
As a result, we experimentally show that such trajectories cannot be directly used to improve learning for path prediction, motivating the present research.

This work proposes AA-SGAN (Adversarially Augmented Social GAN), an end-to-end adversarial approach for predicting trajectories over a combination of real and synthetic trajectories in input.
In a nutshell, we introduce a generative \textit{Augmenter} to manipulate synthetic trajectories.
Such \textit{synth-augmented} trajectories, interleaved with real ones, are input into a Generator that learns to predict a trajectory continuation as in~\cite{gupta2018social}.
Notably, the whole architecture is trained end-to-end propagating the gradient of an adversarial loss from the Discriminator back towards the Generator and the Augmenter. As a result, the Generator improves its prediction accuracy while at the same time the Augmenter learns how to increase the diversity of synthetic trajectories. As a qualitative example, the Augmenter modifies the synthetic trajectories shown in Fig.~\ref{fig:syntplot} to those reported in Fig.~\ref{fig:ganplot} that bear more resemblance to real ones.
Our experiments over real test sets~\cite{eth2009, ucy2007} confirm that our method yields significant gains over a reference SGAN architecture trained either over real trajectories only or a hybrid of real and synthetic trajectories.

The paper is organized as follows: Section ~\ref{sec:related_works} introduces relevant background. Section \ref{sec:method} illustrates our proposed methodology, while Section \ref{sec:experiments} presents our experimental evidence.
Finally, Section \ref{sec:conclusions} draws the conclusions of our work.

\section{Background and Related Works}
\label{sec:related_works}

In this section, we present the background relevant to the understanding of this work.
Namely, we discuss existing approaches to pedestrian path prediction and the datasets most commonly used for this task.

\subsection{Trajectory Prediction Methods}

Over the years, pedestrian trajectory prediction has been the subject of many endeavours~\cite{pedestrian_review}.
Early attempts tried to model this complex task with models borrowed from classic Physics \cite{Helbing_1995, Pellegrini2010, TPAMI.2015.2430335}. Recently, however, learning-based models have outperformed such early approaches and represent the state of the art in the field.
Therefore, our literature review will be limited to learning-based approaches.
While some based methods rely on multimodal inputs (e.g., video beside trajectories) to boost performance, the present work relies on unimodal trajectories-based inputs. Our review will also be constrained to such cases.

Given the sequential nature of predicting a future trajectory based on past observations, Recurrent Neural Networks (RNNs) based methods were among the first to yield promising results among learning-based methods.
Namely,~\cite{alahi2016} proposes to rely on Long Short Term Memory (LSTM) RNNs enhanced by an ad-hoc \textit{social pooling} layer.
This layer models the interactions between nearby pedestrians and is responsible for guaranteeing that the predicted trajectories are also plausible from a social perspective (e.g., paths should not interfere at the same time, social conventions such as keeping the right way should be respected, etc.).
Although modern architectures tend to outperform the aforementioned one, some of its key concepts have been borrowed and utilized in subsequent adversarial learning-based approaches.

\begin{figure*}[ht]
    \centering
    \includegraphics[width=0.8\linewidth]{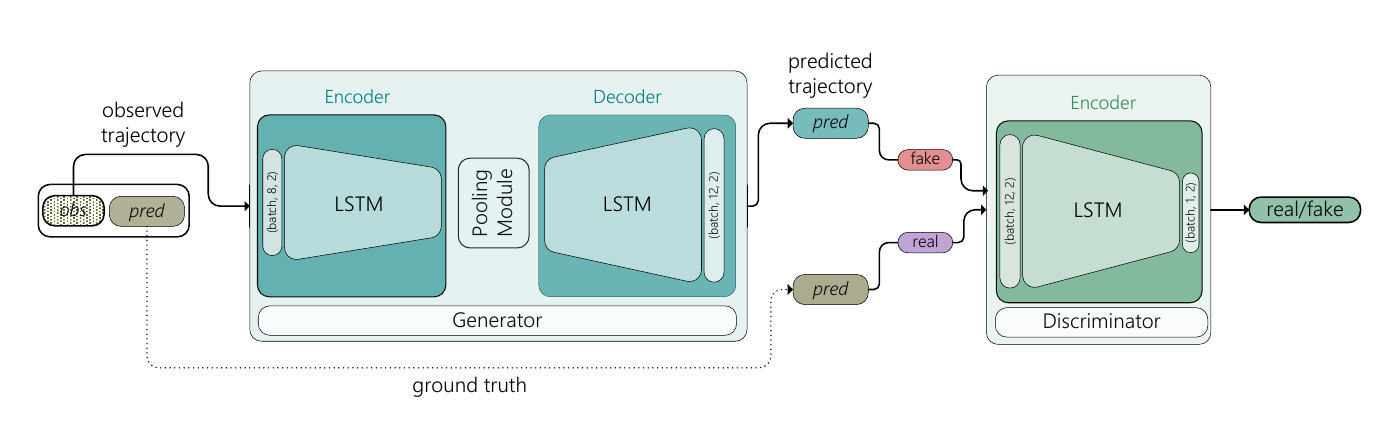}
    \caption{\textit{Social GAN} for pedestrian trajectories prediction~\cite{gupta2018social} adversarial architecture. The Generator learns to generate real-looking trajectories, the Discriminator learns to tell real from generated trajectories. A social pooling module and an ad-hoc loss function enforce the social plausibility of generated trajectories.}
    \label{fig:sgan}
\end{figure*}

Social GAN (shortly, \textit{SGAN}) \cite{gupta2018social} approaches trajectory prediction with an adversarial approach~\cite{goodfellow2014generative}.
Figure~\ref{fig:sgan} shows the internals of the SGAN: the \textit{Generator} receives in input the first 8 steps (observed trajectory, "\textit{obs}") of a 20 steps long pedestrian trajectory and predicts the next 12 steps (predicted trajectory, "\textit{pred}") of the sequence.
Within the Generator, both the encoder and the decoder are implemented as recurrent LSTM networks, connected by a social pooling layer.
The \textit{Discriminator} receives in input the 12 steps sequence output by the Generator and the corresponding ground truth sub-sequence of the same length.
SGAN is trained with an adversarial approach where the Generator learns to produce real-looking path traces, whereas the Discriminator learns to tell generated from real trajectories.
The architecture is trained to minimize a novel variety loss that encourages diversity among generated predictions. Other works using GAN \cite{sophie19} or synthetic data have been proposed \cite{liang2020simaug, liang2020garden}. However, these also integrate visual information into their decision pipeline. We will not deal with these works as our approach is based on using trajectories alone as input to the model.
This approach outperforms prior work in terms of accuracy, variety, collision avoidance, and computational complexity.

\subsection{Real-World Datasets}
\label{sec:realdatasets}

Common datasets used for pedestrian trajectory prediction include ETH \cite{eth2009}, and UCY \cite{ucy2007}. These datasets consist of trajectories extracted from surveillance camera videos and annotated every 0.4 seconds.
Each dataset includes multiple trajectories and each trajectory includes multiple pedestrians.
Each dataset sample includes a temporal index and, for each pedestrian, the pedestrian's identification code and its position in an \textit{(x,y)}-plane.
The ETH dataset includes two different environments called \texttt{biwi\_eth} and \texttt{biwi\_hotel}, whereas the UCY dataset includes five different environments called \texttt{crowds\_zara01}, \texttt{crowds\_zara02}, \texttt{crowds\_zara03}, \texttt{students001}, \texttt{students003} and \texttt{uni\_examples}.
These datasets include a total of 2,205 frames and 6,441 pedestrians. 
These datasets include crowded environments with challenging scenarios such as group behaviour, people crossing each other, avoiding collisions, and groups forming and dispersing.
Therefore, these datasets are commonly employed for researching pedestrian trajectory prediction.

\subsection{Data augmentation and synthetic datasets}
Classical data augmentation, which is based on the generation of additional data, has become quite standard for visual data where manually designed transformations, e.g. crop, colour jitter, rotation, etc., can be applied to images in order to increase the variability in acquisition settings, thus promoting better generalization on real data. 
GANs have recently attracted lots of attention in order to overcome the limits of manual augmentation through the direct synthesis of new images. Nonetheless, GAN models need to be trained on real data as well, thus limiting their applicability in many cases.
Other works have focused on automating data augmentation policies: see as an example \cite{zhang2020adversarial} and reference therein.

Data augmentation for path prediction can be very critical: also the authors in \cite{gupta2018social} reported that synthetic data could potentially lead to worse performance.
Therefore, the use of simulated trajectories in this context, while very promising, remains an open issue.
One attempt is based on creating a synthetic dataset starting from real trajectories. This is achieved by randomly sampling a trajectory from real data, adding a small perturbation with a translation, reverting the path by flipping the starting and ending points, and truncating a random number of steps. This process creates synthetic data highly dependent on real data that preserves many of its characteristics, making the crafted trajectories not suitable for proper augmentation \cite{DBLP:journals/corr/abs-1903-01860}.
Another work in this direction is based on modelling the underlying physiological, and psychological factors that affect pedestrian movement with agents \cite{10.1371/journal.pone.0117856}. This attempts to develop an algorithm based on density-dependent filters to generate human-like crowd flows. The approach borrows deeply from the physical model based on reciprocal velocity obstacles and social forces to create synthetic data.
Promising results were obtained using the synthetic dataset for visual tasks, e.g. pedestrian tracking \cite{fabbri21iccv}. 
Extending this approach to path prediction can be very critical: synthetic trajectories are likely to be simplified and too predictable, e.g., due to game engine scripting. Although they are suitable for learning simple path prediction models, they fail to mimic human behaviour correctly. Synthetic trajectories poorly represent avoidance paths that people typically and unconsciously adopt when walking with others around. 
To overcome such limitations in this work we propose to use synthetic data as input to a new Augmenter module that is trained end-to-end along with the path prediction task.
\section{Proposed Method}
\label{sec:method}

In this section, we describe our proposed approach towards learning to predict pedestrian trajectories from both real and augmented synthetic trajectories.

\subsection{Problem definition}
Let us define a pedestrian \textit{trajectory} as a sequence of $t_{pred}$ samples in the temporal order.
Each sample is a pair of coordinates in space, where each element $(x_{t},y_{t})^{(i)}$ ($i=\{ 1, \ldots, N \}$)
represents the position of the $i$-th pedestrian at time-instant $t \in [1, t_{pred}]$. 
We have that $t =\{1, \ldots, t_{obs} , t_{obs} +1, \ldots, t_{pred}\}$, where $t=t_{obs}$ is the number of \textit{observed} samples and $(t_{pred} - t_{obs})$ is the number of following samples to be predicted.
As a common practice in the related literature, all the trajectory coordinates are preliminarily normalized to relative coordinates with respect to the starting point. 

\subsection{Architecture}

\begin{figure*}[t!]
    \centering
    \includegraphics[width=0.98\linewidth]{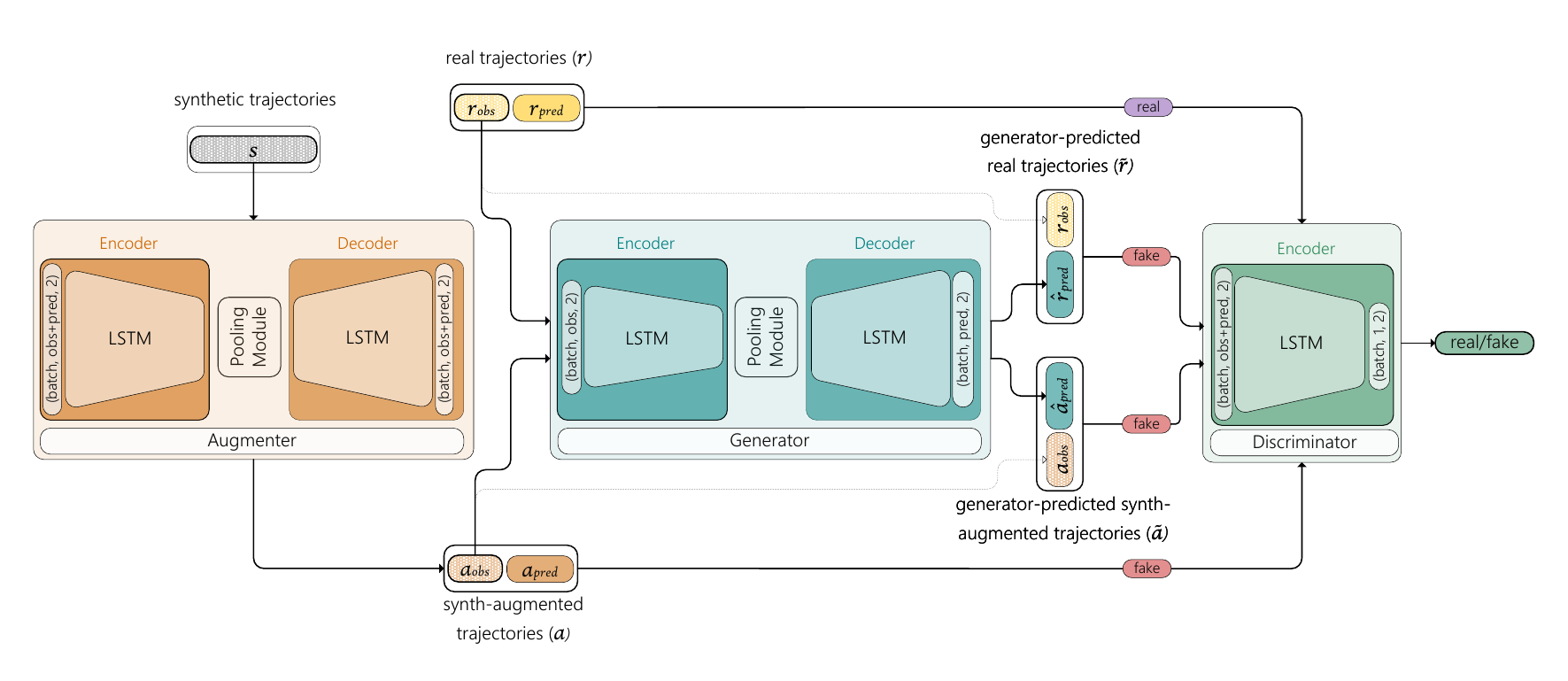}
    \caption{AA-SGAN for adversarially augmented pedestrian trajectories prediction architecture. The Augmenter learns to augment synthetic trajectories into synth-augmented; the Generator learns to generate trajectories prediction; the Discriminator learns to discriminate real from generated and synth-augmented trajectories.}
    \label{fig:workflow}
\end{figure*}

Our proposed AA-SGAN architecture is shown in Fig.~\ref{fig:workflow}.
As for~\cite{gupta2018social}, real trajectories $r$ are fed as input to a Generator $G$, whose task is to predict the future samples.
The Generator receives as input the first $t_{obs}$ samples of a trajectory and predicts the next $(t_{pred}- t_{obs})$ samples.
However, in the proposed architecture, the Generator receives in input also \textit{synth-augmented} trajectories $a$.
Synth-augmented trajectories are generated by an  \textit{Augmenter} $A$ that receives in input a synthetic trajectory $s$ and outputs an augmented trajectory $a$.
Therefore, our Generator learns on a larger variety of inputs than only real trajectories as in~\cite{gupta2018social}.
As for~\cite{gupta2018social}, a Discriminator $D$ attempts to discriminate if a trajectory is real or it is fake.
However, our Discriminator is fed with three different classes of fake trajectories, i.e. it is given a chance to learn over a significantly more challenging problem.
Another peculiarity of our proposal is that the above architecture allows us to train the Augmenter not only end-to-end but also over an adversarial loss, rather than just minimizing the difference between its input and output (e.g., MSE).
Adversarial Augmenter training is the key towards augmented synthetic trajectories that are practically useful for training the Generator, as we experimentally show later on.

\paragraph{Augmenter}
The Augmenter $A$  receives in input synthetic trajectories $s$ of length $t_{pred}$ and outputs \textit{synth-augmented} trajectories $a=A(s)$ of identical length.
In detail, the Augmenter relies on an encoder-decoder architecture. Thus, sequences with length $t_{pred}$ are embedded into a Multi-Layer Perceptron layer $\phi(\cdot)$ with ReLU non-linearity to get fixed length vectors $e^{(i)}_t = e_t, t\in[1, t_{pred}], \forall i$. Subsequently, these embedding vectors $e$ are used as input to the RNN model - in this case, we use the LSTM cell $f(\cdot)$ - of the encoder $\epsilon$ at time $t$ with the following recurrence:
\begin{equation}
    \begin{aligned}
    e_t = \phi(x_t, y_t; W^{(\epsilon)}_{e}) \\
    h^{(\epsilon)}_t = f(h_{t-1}, e_t; W^{(\epsilon)})
    \end{aligned}
\end{equation}
where $W^{(\epsilon)}_{e}$ are the embedding weights and, $W^{(\epsilon)}$ are the encoder weights shared between all people in a scene. 
Once observed trajectories are encoded, we use the Pooling Module $\rho(\cdot)$ proposed by Gupta \textit{et al.}, 2018 \cite{gupta2018social} to model human-human interactions: the idea is to obtain a pooled tensor $p_i=p, \forall i$ - consistent with the past - to initialize the hidden state of the decoder $\delta$. Thus, we embed pedestrians through an MLP layer $\gamma(\cdot)$ with ReLU non-linearity and embedding weights $W_c$. 
\begin{equation}
    \begin{aligned}
    c_t = \gamma(p; W_{c})
    \end{aligned}
\end{equation}
Hence, we initialize the decoder state $h^{(\delta)}_t$ as concatenation of $c_t$ and $z$ from $\mathcal{N}(0,1)$:
\begin{equation}
    \begin{aligned}
    h^{(\delta)}_t = [c_t, z]
    \end{aligned}
\end{equation}
Now, it is possible to generate trajectories for all pedestrians as follows:
\begin{equation}
    \begin{aligned}
    e_t = \phi(x_{t-1}, x_{t-1}; W^{(\delta)}_{e}) \\
    p = \rho(h^{(\delta)}_{t-1}, \dots, h^{(\delta)}_{t}) \\
    h^{(\delta)}_{t} = f(\gamma(p, h^{(\delta)}_{t-1}), e_t; W^{(\delta)}) \\
    \end{aligned}
\end{equation}
where $W^{(\delta)}_{e}$ are the embedding weights and $W^{(\delta)}$ are the decoder weights shared between all people in a scene. Then, the coordinates will be:
\begin{equation}
    (\hat{x}_t, \hat{y}_t) = \gamma(h^{(\delta)}_t)
\end{equation}

\paragraph{Generator}
The Generator $G$ observes the first $t_{obs}$ samples of a trajectory of $t_{pred}$ samples and predicts the next $t_{pred}-t_{obs}$ samples. The architecture of the predictor has an encoder-decoder structure that is entirely analogous to that of the Augmenter $A$.
However, unlike~\cite{gupta2018social}, the Generator takes two types of inputs: real trajectories $r$ and synth-augmented trajectories $a$ output by the Augmenter. 
More in detail, the real trajectory $r$ is divided in two segments called $r_{obs}$ ($t_{obs}$ samples) and $r_{pred}$ ($t_{pred} - t_{obs}$ samples).
Then, $r_{obs}$ is input into $G$ that outputs the prediction $\hat{r}_{pred}=G(r_{obs})$, i.e. the predicted continuation of $r_{obs}$.
Similarly, synth-augmented trajectories are split in $a_{obs}$ and $a_{pred}$. The Generator is similarly fed with $a_{obs}$ and predicts $\hat{a}_{pred} = G(a_{obs})$.
The best trajectory is selected by calculating the $L2$-distance for each predicted point, following the method proposed by \cite{gupta2018social}.
Predictions $\hat{r}_{pred}$ and $\hat{a}_{pred}$ are then concatenated with ${r}_{obs}$ and ${a}_{obs}$ as $\hat{r}$ and $\hat{a}$, respectively.

\paragraph{Discriminator}
The Discriminator $D$ consists of a separate encoder with an MLP layer as the last layer to classify if a trajectory is fake or not, i.e., whether a trajectory is socially acceptable. To this end, $D$ is designed as a binary classifier that takes in input trajectories of $t_{pred}$ samples.

In our proposed architecture, the Discriminator has two peculiarities over~\cite{gupta2018social}.
The first peculiarity is that  the Discriminator is challenged to discriminate between true or fake trajectories over four different classes of inputs rather than just two.
Namely, real trajectories $r$ are the only ones that are labelled as real.
Generator-predicted real trajectories $\tilde{r}=[r_{obs}, \hat{r}_{pred}]=[r_{obs}, G(r_{obs})]$, Generator-predicted synth-augmented trajectories $\tilde{a}=[a_{obs}, G(a_{obs})] = [a_{obs}, \hat{a}_{pred}]$
and synth-augmented trajectories $a$ that go all under the fake label.

This increased variety of false trajectories is expected to provide the Discriminator with more challenging examples at training time.
In turn, the Discriminator will propagate better error gradients when the entire architecture is trained as follows.
Secondly, the Discriminator receives the entire trajectories as input: in fact, the trajectories predicted by the Generator are concatenated with their observed part ($\tilde{r}$ and $\tilde{a}$) and the real trajectories $r$ and synth-.augmented $a$ are taken in their entirety. In this way, it is possible to use the same Discriminator on both the Generator output and the Augmenter output.

\subsection{Training Procedure}

The above-mentioned is trained end-to-end with an adversarial approach~\cite{goodfellow2014generative}.

The Discriminator is trained with the following loss function: 
\begin{equation}
    \begin{aligned}
        L_{D} = \mathbb{E}[ \log(D(r)) + \log( 1 - D(a)) + \\ \log( 1 - D(\tilde{r}))+\log( 1 - D(\tilde{a}))]
    \end{aligned}
    \label{eqn:LD}
\end{equation}
\noindent
with the aim of maximizing the average of the log probability of real trajectories $r$ and the log of the inverse probability for synth-augmented $a$, Generator-predicted real $\tilde{r}$ and Generator-predicted synth-augmented $\tilde{a}$ trajectories.

\noindent
\\
Concerning the Augmenter, an $L2$-loss is computed between $s$ and $a$ overall $t_{pred}$ samples, as proposed in \cite{gupta2018social}.
Thus, the Augmenter is trained to minimize the loss:
\begin{equation}
    \begin{aligned}
        L_{A} = \mathbb{E}[\log(D(a)) + L2(s, a)]
    \end{aligned}
    \label{eqn:lossA}
\end{equation}
\noindent
Indeed, it may seem counterintuitive to introduce an $L2$-loss term when training the Augmenter (we want to make these trajectories different, not identical).
However, we observed that, in combination with the Discriminator loss, this yields trajectories that are more useful to train the Generator
\\
Finally, concerning the Generator, two $L2$-losses are computed between $\tilde{r}$ and $r$ and between $\tilde{a}$ and $a$.
However, the Generator shall be obviously trained only over the $t_{pred} - t_{obs}$ predicted samples.
Thus, the Generator is trained to minimize the loss

\begin{equation}
    \begin{aligned}
        L_{G} = \mathbb{E}[ \log(D(\tilde{r})) + L2(r_{pred}, \hat{r}_{pred}) + \\  \log(D(\tilde{a})) + L2(a_{pred}, \hat{a}_{pred})]
    \end{aligned}
\end{equation}
\noindent

The above described architecture is trained with the classical GAN training procedure, yet extended to the Augmenter. As a first step, the Discriminator $D$ is first optimised over a real trajectory $r$, then over the Generator-predicted trajectory $\tilde{r}$, next over synth-augmented trajectory $a$ and finally over the Generator-predicted synth-augmented trajectory $\tilde{a}$. As a second step, the Generator $G$ is optimised first over real trajectories $r$ and then over the synthesised trajectories. As a final step, the Augmenter $A$ is optimised through the synthetic trajectories$s$.

\section{Experimental results}
\label{sec:experiments}

\begin{table*}
    \centering
    \resizebox{0.9\textwidth}{!}{%
        \begin{tabular}{cl|ccc|c}
        \multicolumn{1}{c|}{\multirow{2}{*}{Metric}} & \multicolumn{1}{c|}{\multirow{2}{*}{Dataset}} & \multicolumn{3}{c|}{\textbf{SGAN} \cite{gupta2018social}} & \textbf{AA-SGAN} \\
        \multicolumn{1}{c|}{}             & \multicolumn{1}{c|}{} & \multicolumn{1}{c|}{real} & \multicolumn{1}{c|}{synthetic} & hybrid &         \\ \hline
        \multicolumn{1}{c|}{}             & \textbf{ETH}          & \multicolumn{1}{c|}{0.85 \footnotesize ($\pm\,\,9\times10^{-3}$)} & \multicolumn{1}{c|}{1.28 \footnotesize ($\pm\,\,1\times10^{-2}$)}      & {0.93  \footnotesize ($\pm\,\,4\times10^{-3}$)}  & {\textbf{0.71} \footnotesize ($\pm\,\,1\times10^{-2}$)} \\ \cline{3-6} 
        \multicolumn{1}{c|}{}             & \textbf{HOTEL}        & \multicolumn{1}{c|}{0.63 \footnotesize ($\pm\,\,4\times10^{-3}$)} & \multicolumn{1}{c|}{0.88 \footnotesize ($\pm\,\,4\times10^{-3}$)}      & {0.63 \footnotesize ($\pm\,\,3\times10^{-3}$)}   & {\textbf{0.42} \footnotesize ($\pm\,\,3\times10^{-3}$)}\\ \cline{3-6} 
        \multicolumn{1}{c|}{ADE} & \textbf{UNIV}         & \multicolumn{1}{c|}{0.67  \footnotesize ($\pm\,\,4\times10^{-4}$)} & \multicolumn{1}{c|}{1.19 \footnotesize ($\pm\,\,8\times10^{-4}$)}      & {0.62  \footnotesize ($\pm\,\,4\times10^{-4}$)}   & \textbf{0.60 \footnotesize ($\pm\,\,8\times10^{-4}$)} \\ \cline{3-6} 
        \multicolumn{1}{c|}{}             & \textbf{ZARA1}        & \multicolumn{1}{c|}{0.42 \footnotesize ($\pm\,\,1\times10^{-3}$)} & \multicolumn{1}{c|}{1.22 \footnotesize ($\pm\,\,2\times10^{-3}$)}      & {0.41 \footnotesize ($\pm\,\,2\times10^{-3}$)}  & \textbf{0.34 \footnotesize ($\pm\,\,2\times10^{-3}$)} \\ \cline{3-6} 
        \multicolumn{1}{c|}{}             & \textbf{ZARA2}        & \multicolumn{1}{c|}{0.40 \footnotesize ($\pm\,\,6\times10^{-4}$)} & \multicolumn{1}{c|}{0.51 \footnotesize ($\pm\,\,8\times10^{-4}$)}      & {0.42 \footnotesize ($\pm\,\,5\times10^{-4}$)}   & \textbf{0.34 \footnotesize ($\pm\,\,7\times10^{-4}$)} \\ \hline
        \multicolumn{2}{c|}{\textbf{Average}}                     & \multicolumn{1}{c|}{0.60} & \multicolumn{1}{c|}{1.02}      & 0.60   & \textbf{0.48} \\ \hline\hline
        \multicolumn{1}{c|}{}             & \textbf{ETH}          & \multicolumn{1}{c|}{1.63 \footnotesize ($\pm\,\,2\times10^{-3}$)} & \multicolumn{1}{c|}{2.58 \footnotesize ($\pm\,\,2\times10^{-2}$)}      & {1.79 \footnotesize ($\pm\,\,8\times10^{-3}$)}   & \textbf{1.24 \footnotesize ($\pm\,\,3\times10^{-2}$)} \\ \cline{3-6} 
        \multicolumn{1}{c|}{}             & \textbf{HOTEL}        & \multicolumn{1}{c|}{1.36 \footnotesize ($\pm\,\,4\times10^{-3}$)} & \multicolumn{1}{c|}{1.79 \footnotesize ($\pm\,\,1\times10^{-2}$)}      & {1.30 \footnotesize ($\pm\,\,6\times10^{-3}$)}   & {\textbf{0.78}  \footnotesize ($\pm\,\,6\times10^{-3}$)} \\ \cline{3-6} 
        \multicolumn{1}{c|}{FDE} & \textbf{UNIV}         & \multicolumn{1}{c|}{1.44 \footnotesize ($\pm\,\,9\times10^{-3}$)} & \multicolumn{1}{c|}{2.36 \footnotesize ($\pm\,\,2\times10^{-3}$)}      & {1.33 \footnotesize ($\pm\,\,9\times10^{-4}$)}   & {\textbf{1.25} \footnotesize ($\pm\,\,2\times10^{-3}$)} \\ \cline{3-6} 
        \multicolumn{1}{c|}{}             & \textbf{ZARA1}        & \multicolumn{1}{c|}{0.90 \footnotesize ($\pm\,\,4\times10^{-3}$)} & \multicolumn{1}{c|}{2.55 \footnotesize ($\pm\,\,6\times10^{-3}$)}      & {0.86 \footnotesize ($\pm\,\,3\times10^{-3}$)}   & {\textbf{0.68} \footnotesize ($\pm\,\,3\times10^{-3}$)} \\ \cline{3-6} 
        \multicolumn{1}{c|}{}             & \textbf{ZARA2}        & \multicolumn{1}{c|}{0.91 \footnotesize ($\pm\,\,2\times10^{-3}$)} & \multicolumn{1}{c|}{1.06 \footnotesize ($\pm\,\,2\times10^{-3}$)}      & {0.93 \footnotesize ($\pm\,\,2\times10^{-3}$)}   & {\textbf{0.71} \footnotesize ($\pm\,\,2\times10^{-3}$)} \\ \hline
        \multicolumn{2}{c|}{\textbf{Average}}                     & \multicolumn{1}{c|}{1.24} & \multicolumn{1}{c|}{2.07}      & 1.24   & \textbf{0.93} \\ \hline\hline
        \end{tabular}
   }
    \caption{Trajectory prediction accuracy in terms of ADE and FDE for a reference SGAN baseline trained on real, synthetic and hybrid (50\% real/ 50\% synthetic) dataset compared with AA-SGAN.}
    \label{tab:results}
\end{table*}

In this section, we evaluate the performance of our AA-SGAN architecture on the two publicly available real datasets ETH and UCY introduced in Sect.~\ref{sec:realdatasets}. 

The experimental evaluation is based on the leave-one-out methodology used in the related literature.
In particular, we consider a total of 5 sets of real trajectories (\texttt{eth}, \texttt{hotel} from ETH and \texttt{univ}, \texttt{zara1} and \texttt{zara2} from UCY dataset): we train on 4 sets and test on the remaining (left out) set. 

Other common settings are $t_{obs}=8$ (3.2 s) and $t_{pred}=20$ (8 s) which amounts to predicting 4.8 s of the future path followed by each pedestrian 
(all possible 8 s long trajectories are extracted from datasets using a sliding window with skip equal to 1 frame as in  \cite{alahi2016, gupta2018social, sophie19}).

\begin{figure}[!tbp]
  \centering
  \includegraphics[width=\linewidth]{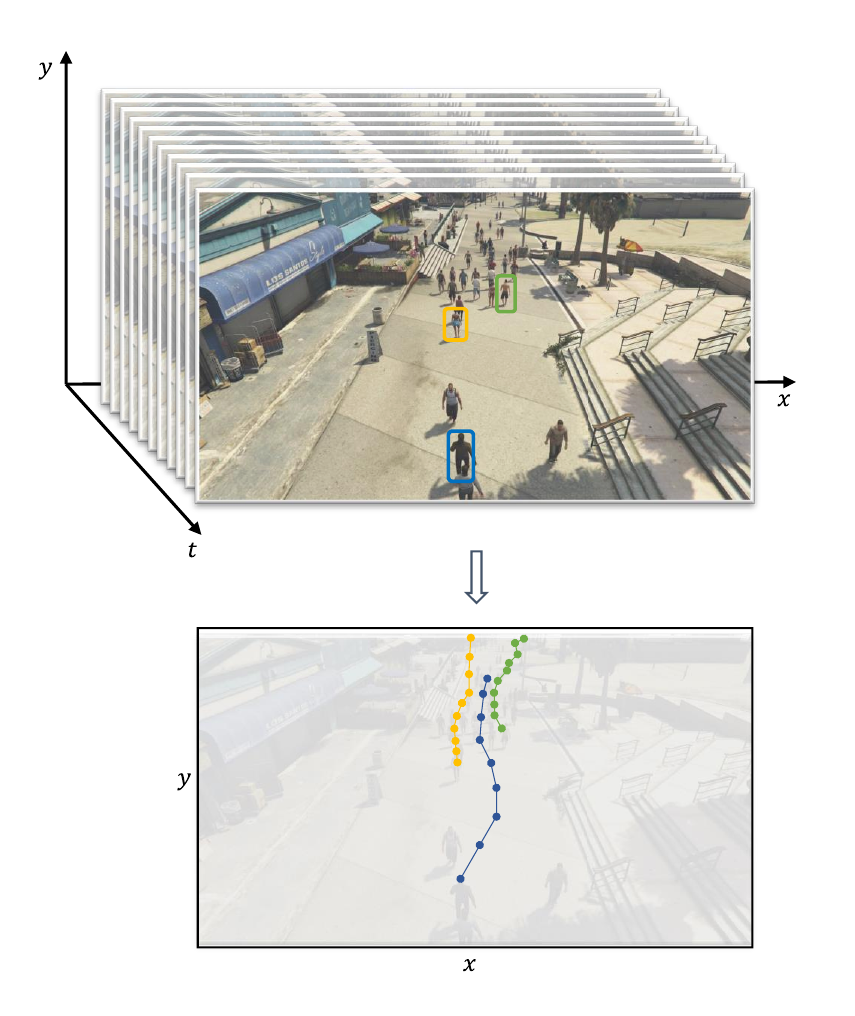}
  \caption{Examples of frames from the JTA dataset (top) and the corresponding trajectories (bottom) we used as a synthetic training set.}
  \label{fig:jta_example}
\end{figure}

The trajectory prediction accuracy is measured in terms of Average Displacement Error (ADE) and Final Displacement Error (FDE) \cite{chandra2019robusttp, gupta2018social, goodfellow2014generative, lee2017desire}. Therefore, given a generic trajectory $\nu$, we have:

\begin{equation}
    ADE_{\nu_{pred}, \hat{\nu}_{pred}}= \sum^{t_{pred}}_{t=t_{obs}+1} \frac{ \|\hat{\nu}_{t}  - \nu_{t}\| }{(t_{pred}-t_{obs})}, \forall i
\end{equation}
        
\begin{equation}
    FDE_{\nu_{pred}, \hat{\nu}_{pred}} = \|\hat{\nu}_{t_{pred}} - \nu_{t_{pred}}\|, \forall i
\end{equation}

where $\nu_{pred}$ and $\hat{\nu}_{pred}$ are, respectively, the ground truth and the model prediction. Moreover, we used the same evaluation approach as SGAN. We evaluated the best-of-N via repeated sampling, with N=20.
\\
As a source of synthetic trajectories for AA-SGAN Augmenter, we use the JTA Dataset \cite{fabbri2018learning}: it consists of a vast dataset of trajectories extracted from the videogame Grand Theft Auto V (see sample frames in Fig.~\ref{fig:jta_example}), containing 384 full-HD videos of 30 seconds and recorded at 30 fps. Different environments  (e.g., airports, stations, squares, and parks), different climatic conditions (e.g., sun and rain), different angles (e.g., from above, from human height, and in motion), and other lighting conditions (e.g., day, night, artificial light) characterize each video. These characteristics make this dataset highly versatile and usable for many purposes.
One of the main advantages of synthetic data is the ability to quickly and easily retrieve a large amount of data and related labels.
To match the structure of the real pedestrian trajectories used for testing, we rely only on JTA trajectories characterized by a top view.
In detail, we track the \textit{head top} skeleton joint to get pedestrian trajectories; also, we maintain the same inter-sample time interval equal to 0.4 s.
Overall, the synthetic dataset includes a total of 4,488 frames from 3,834 pedestrians.

\subsection{Path Prediction Accuracy}
\label{sec:results}
Table~\ref{tab:results} shows the trajectory prediction accuracy obtained by our proposed AA-SGAN scheme and the three baselines SGAN schemes.
The \textit{ AA-SGAN} scheme refers to the proposed architecture trained as described above.
The three baseline schemes refer to a standard SGAN trained, in turn, as follows.
The \textit{real} scheme corresponds to the setup~\cite{gupta2018social}, where an SGAN is trained over real trajectories from UCY and ETH.
The \textit{synthetic} scheme corresponds to the case where an SGAN is trained only over synthetic trajectories from JTA.
Finally, \textit{hybrid} refers to the case where SGAN is trained over a 50-50 mix of real and synthetic trajectories from JTA (the size of this hybrid dataset is twice that of the former two).
Notice that this mix of real and synthetic trajectories is the same provided in input to our AA-SGAN scheme, as discussed later on.
The four schemes share the same training hyper-parameters and configuration suggested in \cite{gupta2018social}.
\\
The results for the \textit{real} scheme reflect those reported in~\cite{gupta2018social}, apart from some minor differences on the \texttt{zara1} and \texttt{zara2}. Please also note that all the results reported in the following are averaged on 3 trials (corresponding standard deviation is shown as well).
As expected, performance drops when SGAN is trained over synthetic data only. This performance loss shows that JTA synthetic trajectories are so simple that they are not useful to train an SGAN.
As an example, on the ETH test, the value for ADE increases from 0.85 up to 1.28 when training on synthetic trajectories only.
The \textit{hybrid} column in Tab.~\ref{tab:results} shows that accuracy does not improve over the \textit{real} scheme. For the ETH experiment, we even report ADE equal to 0.93 which is significantly worse than \textit{real}.
These preliminary experiments show that synthetic trajectories cannot replace real ones at training times nor do they bring any benefit if mixed with them.
\\
The last column shows that the proposed AA-SGAN method yields much better accuracy in all cases, both in terms of ADE (20\% reduction from 0.6 to 0.48) and FDE (25\% reduction from 1.24 to 0.93).
We recall that the AA-SGAN scheme is trained over the same mix of real and synthetic trajectories used for the \textit{hybrid} scheme.
Such difference in accuracy despite the same training set can be brought down to the job performed by the Augmenter, which makes synthetic trajectories eventually useful for training.
In the AA-SGAN scheme, we used a 50-50 ratio of real and synthetic trajectories: this is a reasonable choice as the Augmenter and Generator are fed with a balanced mix. 

In the following ablation study, we will discuss whether it is useful to alter such a real-synthetic ratio and to what extent.

\begin{figure}[!tbp]
    \centering
        \begin{subfigure}[h]{\linewidth}
            \centering
            \includegraphics[width=0.8\linewidth]{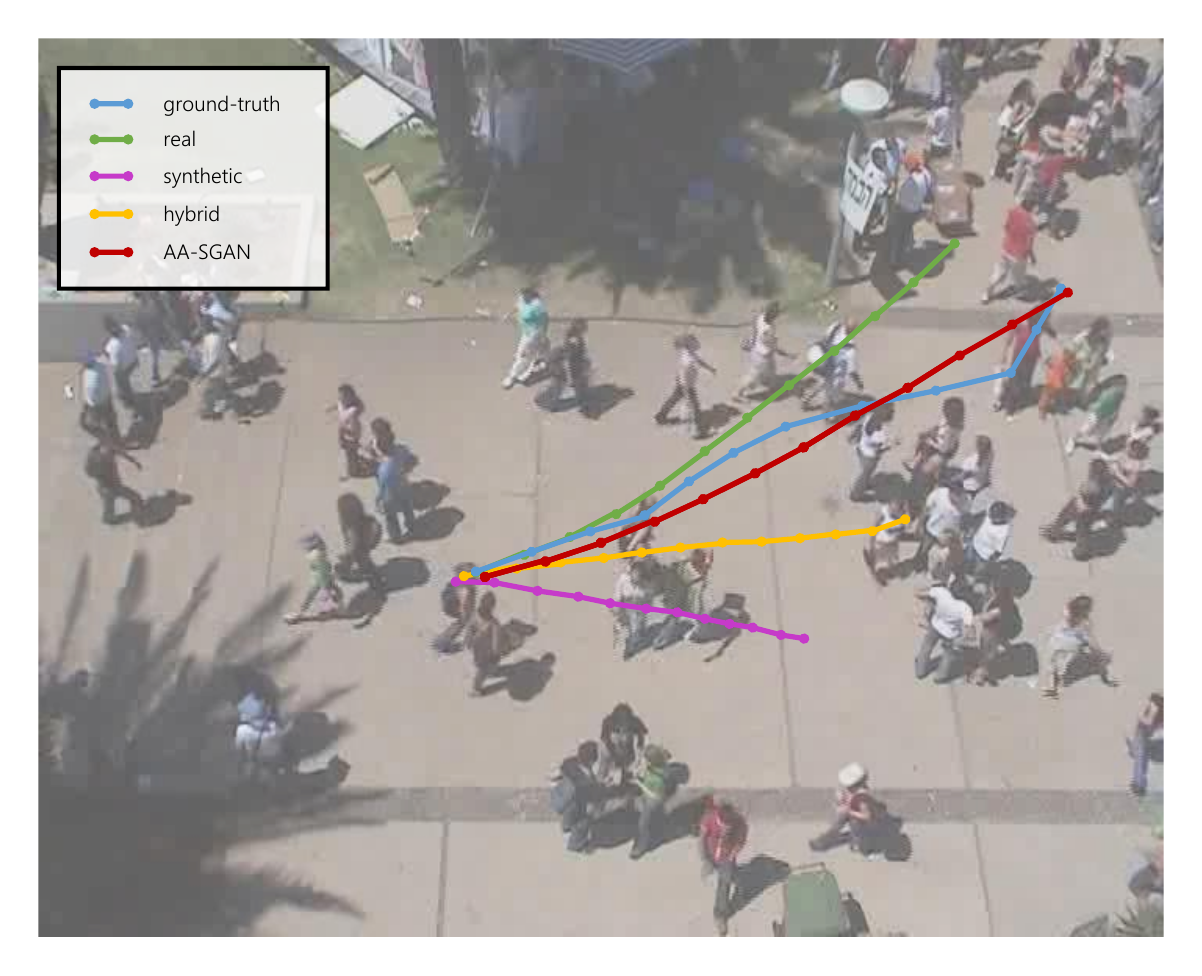}
            \caption{\texttt{univ}}
            \label{fig:dataset_examples_a}
        \end{subfigure}
            \hspace{0.01\linewidth}
        \begin{subfigure}[h]{\linewidth}
            \centering
            \includegraphics[width=0.8\linewidth]{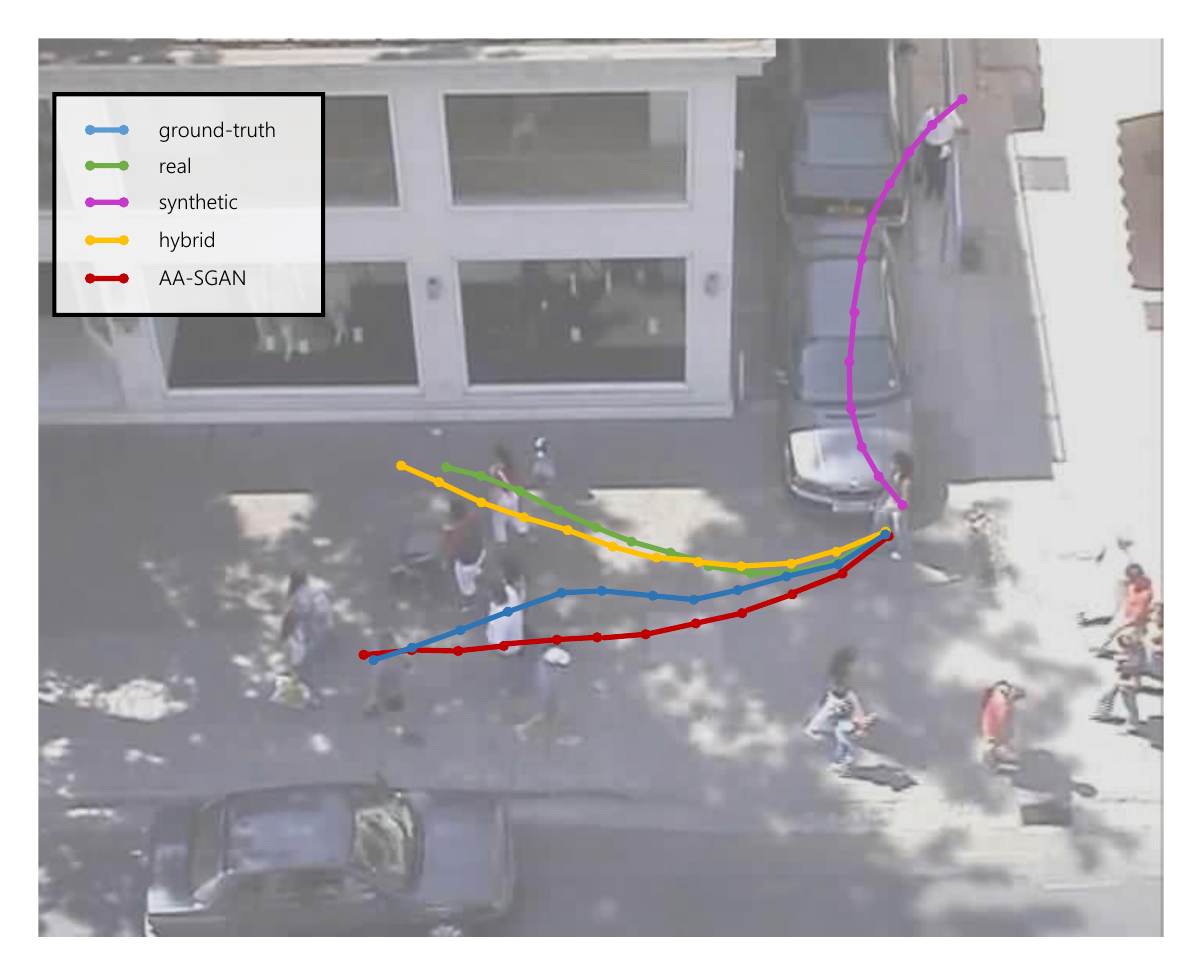}
            \caption{\texttt{zara1}}
            \label{fig:dataset_examples_b}
        \end{subfigure}
    \caption{Visual comparison between the groud-truth (blue) and predicted trajectory by SGAN with real (green), synthetic (magenta) and hybrid (yellow) training. Predictions by the proposed AA-SGAN are in red and are the closest to the ground truth. The selected results are taken from \texttt{univ} and \texttt{zara1} testset.}
    \label{fig:dataset_examples}
\end{figure}

Figure \ref{fig:dataset_examples} illustrates some trajectories predicted by the four schemes in Table \ref{tab:results}. The \textit{synth} scheme predicted trajectories are the worst, diverging the most from the ground truth other than being not acceptable due to collisions (see Fig \ref{fig:dataset_examples_b}).
On the contrary, the results of the AA-SGAN scheme are the closest to the ground truth.

\subsection{Ablation Study}

Before moving further with the experiments, we recall that the experiments with schemes \textit{real}, \textit{synthetic} and \textit{hybrid} in Tab.~\ref{tab:results} above can be already interpreted as an ablation study.
In fact, the standard SGAN architecture trained over a hybrid dataset can be seen as equivalent to AA-SGAN minus the Augmenter, where synthetic trajectories are fed directly into the Generator. 

\paragraph{Independent Augmenter training} In this first ablation experiment, we investigate the advantage of \textit{jointly} training the Augmenter $A$ and the Generator $G$ in an adversarial framework.
We recall that in our architecture, $G$ and $A$ are jointly optimized over the gradient of the adversarial loss function backpropagated by the Discriminator $D$. In order to investigate on this advantage, we removed $G$ from the AA-SGAN architecture and we train the resulting GAN, between $A$ and $D$, using their corresponding losses.
As a consequence, in this ablated architecture the Augmenter will create synth-augmented
trajectories without any feedback from $G$, i.e. without being able to appreciate their contribution to solving the prediction task; the obtained  synth-augmented are stored offline for the subsequent training.
Next, these trajectories are used to train a reference SGAN as in the \textit{hybrid} scheme (50\% real and 50\% synth-augmented trajectories). $G$ and $D$ are trained according to the same adversary loss functions described in the previous section.

Tab.~\ref{tab:end2end-vs-offline} compares the results of this scheme with AA-SGAN.
Performance is still above the \textit{real} baseline, however, it is well below AA-SGAN.
This experiment confirms the importance of jointly training the Augmenter $A$ with the rest of the architecture.

\begin{table}
    \centering
    \begin{tabular}{cc|c|c}
    \multicolumn{1}{c|}{Metric}      & Dataset & \textbf{AA-SGAN} & \textbf{\begin{tabular}[c]{@{}c@{}} Independent\\Augmenter \end{tabular}} \\ \hline
    \multicolumn{1}{c|}{\multirow{5}{*}{ADE}} & \textbf{ETH}     & \textbf{0.71}    & 0.74                                                                              \\ \cline{3-4} 
    \multicolumn{1}{c|}{}                     & \textbf{HOTEL}   & \textbf{0.42}    & 0.57                                                                              \\ \cline{3-4} 
    \multicolumn{1}{c|}{}                     & \textbf{UNIV}    & \textbf{0.60}    & 0.61                                                                              \\ \cline{3-4} 
    \multicolumn{1}{c|}{}                     & \textbf{ZARA1}   & \textbf{0.34}    & 0.36                                                                              \\ \cline{3-4} 
    \multicolumn{1}{c|}{}                     & \textbf{ZARA2}   & \textbf{0.34}    & 0.37                                                                              \\ \hline
    \multicolumn{2}{c|}{\textbf{AVG}}                            & \textbf{0.48}    & 0.53                                                                              \\ \hline \hline
    \multicolumn{1}{c|}{\multirow{5}{*}{FDE}} & \textbf{ETH}     & \textbf{1.24}    & 1.45                                                                              \\ \cline{3-4} 
    \multicolumn{1}{c|}{}                     & \textbf{HOTEL}   & \textbf{0.78}    & 1.09                                                                              \\ \cline{3-4} 
    \multicolumn{1}{c|}{}                     & \textbf{UNIV}    & \textbf{1.25}    & 1.30                                                                              \\ \cline{3-4} 
    \multicolumn{1}{c|}{}                     & \textbf{ZARA1}   & \textbf{0.68}    & 0.75                                                                              \\ \cline{3-4} 
    \multicolumn{1}{c|}{}                     & \textbf{ZARA2}   & \textbf{0.71}    & 0.78                                                                              \\ \hline
    \multicolumn{2}{c|}{\textbf{AVG}}                            & \textbf{0.93}    & 1.07                                                                              \\ \hline \hline
    \end{tabular}
    \caption{Trajectory prediction accuracy of AA-SGAN (joint Augmenter training) versus independent training of the Augmenter feeding reference SGAN.}
    \label{tab:end2end-vs-offline}
\end{table}

\paragraph{Synthetic to real ratio} Another aspect worth being investigated is the ratio between the real and the synthetic trajectories used to train AA-SGAN.
Tab.~\ref{tab:synth-vs-allsynth} shows the prediction accuracy when the synthetic trajectories increase by a 10-fold factor (the number of real trajectories remains constant).
On average, a drop of about 10\% in ADE and FDE is observed, showing the importance of balancing real and synthetic trajectories.
We hypothesize that this predominance of synthetic traces makes adversarial training less stable, explaining the performance drop.

\begin{table}
    \centering
    \begin{tabular}{cc|c|c}
     &  & \multicolumn{2}{c}{Real-Synthetic ratio} \\ \hline
     Metric      & Dataset & ~~1-to-1~~ & 1-to-10 \\ \hline
    \multicolumn{1}{c|}{\multirow{5}{*}{ADE}} & \textbf{ETH}     & 0.71             & 0.70                                                                   \\ \cline{3-4} 
    \multicolumn{1}{c|}{}                     & \textbf{HOTEL}   & 0.42             & 0.54                                                                   \\ \cline{3-4} 
    \multicolumn{1}{c|}{}                     & \textbf{UNIV}    & 0.60             & 0.69                                                                   \\ \cline{3-4} 
    \multicolumn{1}{c|}{}                     & \textbf{ZARA1}   & 0.34             & 0.35                                                                   \\ \cline{3-4} 
    \multicolumn{1}{c|}{}                     & \textbf{ZARA2}   & 0.34            & 0.34                                                                  \\ \hline
    \multicolumn{2}{c|}{\textbf{AVG}}                            & 0.48             & 0.52                                                                 \\ \hline \hline
    \multicolumn{1}{c|}{\multirow{5}{*}{FDE}} & \textbf{ETH}     & 1.24            & 1.25                                                                   \\ \cline{3-4} 
    \multicolumn{1}{c|}{}                     & \textbf{HOTEL}   & 0.78            & 1.09                                                                   \\ \cline{3-4} 
    \multicolumn{1}{c|}{}                     & \textbf{UNIV}    & 1.25             & 1.38                                                                   \\ \cline{3-4} 
    \multicolumn{1}{c|}{}                     & \textbf{ZARA1}   & 0.68             & 0.70                                                                   \\ \cline{3-4} 
    \multicolumn{1}{c|}{}                     & \textbf{ZARA2}   & 0.71             & 0.68                                                                   \\ \hline
    \multicolumn{2}{c|}{\textbf{AVG}}                            & 0.93            &    1.02                                                                    \\ \hline \hline
    \end{tabular}
    \caption{ADE and FDE values for 1-to-1 and 1-to-10 ratio between real and synthetic (R/S) trajectories used in AA-SGAN training (the number of real trajectories does not change).}
    \label{tab:synth-vs-allsynth}
\end{table}

\newpage
\section{Conclusions}
\label{sec:conclusions}

This work proposed AA-SGAN, a generative architecture for predicting accurate pedestrian trajectories leveraging synthetic trajectories beside real ones.
We experimentally showed that computer-generated synthetic trajectories bring no benefit when used to train the state-of-the-art SGAN generative model.
However, if synthetic trajectories are first augmented before being fed to the Generator (\textit{synth-augmented} trajectories), they boost the diversity of the training set, improving the accuracy of the predictions.
Through an ablation study, we show that joint training of the Augmenter with the rest of the architecture is the first key element towards accurate trajectory predictions.
Through ablation, we also show that a balanced ratio between real and synthetic trajectories is another key element of our architecture.

%
%
\bibliographystyle{splncs04}
\bibliography{main}
\end{document}